\def\year{2016} 
\documentclass[letterpaper]{article} \usepackage{aaai16} \usepackage{times}
\usepackage{helvet} \usepackage{courier} \usepackage{multirow}
\usepackage{graphicx} \usepackage{subfigure} \usepackage{amsfonts}
\usepackage{comment}
\begin{document} 
%
\title{Deep Learning with S-shaped Rectified Linear Activation Units} %
\author{Xiaojie Jin$^1$ Chunyan Xu$^4$ Jiashi Feng$^2$ Yunchao Wei$^2$ Junjun Xiong$^3$ Shuicheng Yan$^2$\\
$^1$NUS Graduate School for Integrative Science and Engineering, NUS\\
$^2$Department of ECE, NUS \qquad $^3$Beijing Samsung Telecom R\&D Center\\
$^4$School of CSE, Nanjing University of Science and Technology\\
\tt\small xiaojie.jin@nus.edu.sg, xuchunyan01@gmail.com, elefjia@nus.edu.sg, \\
\tt\small wychao1987@gmail.com, jjun.xiong@samsung.com, eleyans@nus.edu.sg}
\maketitle
\begin{abstract}
\begin{quote} Rectified linear activation units are important components for
state-of-the-art deep convolutional networks. In this paper, we propose
a novel S-shaped rectified linear activation unit (SReLU) to learn both convex and non-convex functions, imitating the multiple function forms given by the two fundamental laws, namely  the
Webner-Fechner law and the Stevens law,  in psychophysics and neural sciences. Specifically, SReLU consists of three piecewise linear
functions, which are formulated by four learnable parameters. The SReLU is learned jointly with the training of the whole deep network through
back propagation. During the training phase, to initialize SReLU in different
layers, we propose a ``freezing" method to degenerate SReLU into a predefined
leaky rectified linear unit in the initial several training epochs and then adaptively learn the good initial values. SReLU can be universally used in the existing deep networks with negligible additional parameters and computation cost. Experiments with two popular CNN architectures, Network in Network and GoogLeNet on scale-various benchmarks including CIFAR10, CIFAR100, MNIST and ImageNet demonstrate that SReLU achieves remarkable improvement compared to other activation functions.
\end{quote}
\end{abstract}

\begin{figure}[!t] \centering \hspace{-0.8cm}
	\includegraphics[width=8cm]{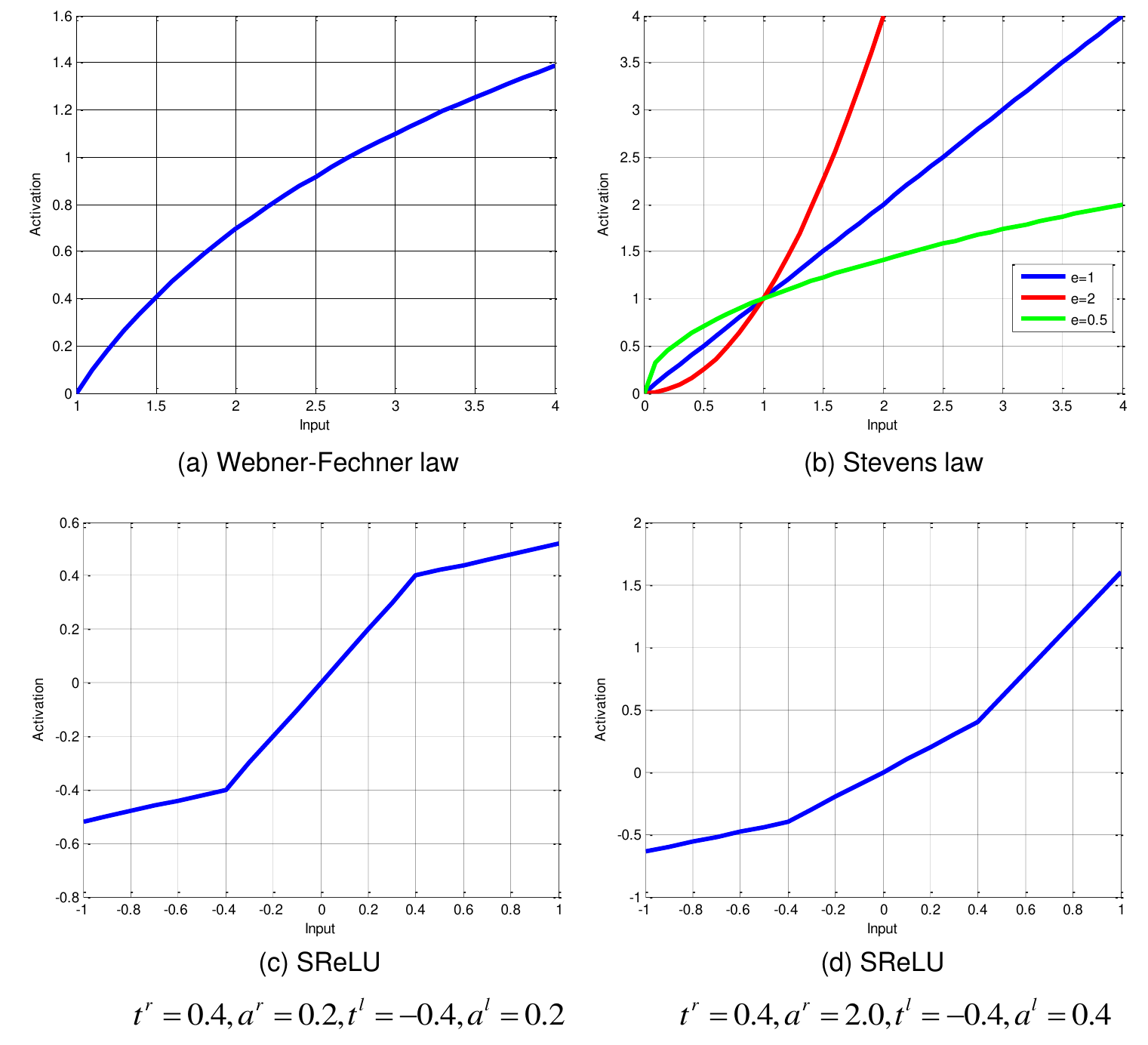}
	\caption{The function forms of Webner-Fechner law and Stevens law along with the proposed
		SReLU.  (a) shows the logarithm function. (b) shows the power function with different exponents. (c) and (d) are different forms of SReLU by changing the
		parameters. The positive part of (c) and (d) are derived by imitating the logarithm
		function (a) and power function (b), respectively. }
	\label{activationlaw}
\end{figure}
\section{Introduction}
\noindent Convolutional neural networks (CNNs) have made great progress in various
fields, such as object classification~\cite{krizhevsky2012imagenet},
detection~\cite{RCNN} and character
recognition~\cite{cirecsan2011convolutional}.  One of the key factors
contributing to the success of the modern deep learning models is using the non-saturated
activation function (\textit{e.g.} ReLU) to replace its saturated counterpart
(\textit{e.g.} sigmoid and tanh), which not only solves the problem of ``exploding/vanishing
gradient" but also makes the deep networks converge fast. Among all the
proposed non-saturated activation functions, the Rectified Linear Unit (ReLU)~\cite{ReLU} is widely viewed as one of the several reasons for the remarkable performance of deep
networks ~\cite{krizhevsky2012imagenet}.

Recently, there are some other activation functions proposed to boost the performance of CNNs. Leaky ReLU (LReLU)~\cite{LReLU}
assigns the negative part with a non-zero slope.~\cite{PReLU} proposed the parametric rectified
linear unit (PReLU), which requires learning the negative part instead of using predefined values. Adaptive piecewise linear activation (APL) proposed in~\cite{learnactivationfunction} sums up several hinge-shared linear
functions.~\cite{maxout} proposed the ``maxout" activation function,
which approximates arbitrary convex functions by computing the maximum of $k$ linear functions for each neuron as the output.

Although the activation functions mentioned above have been reported to achieve
good performance in CNNs, they all suffer from a weaknesses, \textit{i.e.,} their limited
ability to learn non-linear transformation. For example, none of ReLU, LReLU, PReLU and maxout can learn the non-convex
functions since they are essentially all convex functions. Although APL can approximate
non-convex function, it requires the rightmost linear function in all the component
functions to have a unit slope and bias 0, which is an inappropriate
constraint and undermines its representation ability.

Inspired by the fundamental Webner-Fechner law~\cite{weber1851annotationes} and Stevens law~\cite{stevens} in psychophysics and
neural sciences, we propose a novel kind of activation unit, namely the S-shaped
rectified linear unit (SReLU). Two examples of SReLU's function forms are
shown in Figure~\ref{activationlaw}(c)(d). Briefly speaking, both the
Webner-Fechner law and the Stevens law describe the relationship between
the magnitude of a physical stimulus and its perceived intensity or strength~\cite{neuralcoding}. The Webner-Fechner law holds that the perceived magnitude $s$ is
a logarithmic function of the stimulus intensity $p$ multiplied by a modality and a
dimension specific constant $k$. That is,
\begin{equation}
\label{eq:Webner-Fechner} s = k\log p.
\end{equation} And the Stevens law explains the relationship through a power
function, \textit{i.e.},
\begin{equation}
\label{eq:Stevens} s = kp^e,
\end{equation} where all the parameters have the same definitions as in the
Webner-Fechner law, except for an additional parameter $e$ which is an exponent depending on the type of the stimulus. The function forms proposed by the two
laws are shown in Figure~\ref{activationlaw}(a)(b). These laws are usually valid for
general sensory phenomena and can account for many properties of sensory
neurons~\cite{animalphysiology}. More detailed discussions will be
presented in Related Work. 

Roughly, SReLU consists of three piecewise linear functions constrained by
four learnable parameters as shown in Eqn.~(\ref{eq:SReLU}). The usage of SReLU brings two advantages to the deep
network. Firstly, SReLU can learn both
convex and non-convex functions, without imposing any constraints on its learnable  parameters, thus the deep
network with SReLU has a stronger feature learning capability. Secondly, since SReLU
utilizes piecewise linear functions rather than saturated functions, thus it shares the same advantages of the non-saturated
activation functions: it does not suffer from the ``exploding/vanishing gradient" problem
and has a high computational speed during the forward and back-propagation of deep
networks. To verify the effectiveness of SReLU, we test it with two popular deep
architectures, Network in Network and GoogLeNet, on
four datasets with different scales, including
CIFAR10, CIFAR100 and MNIST and ImageNet. The experimental
results have shown remarkable improvement over other activation functions.

\section{Related Work}
\begin{figure}[!t] \centering \hspace{-0.2cm}
\includegraphics[width=8cm]{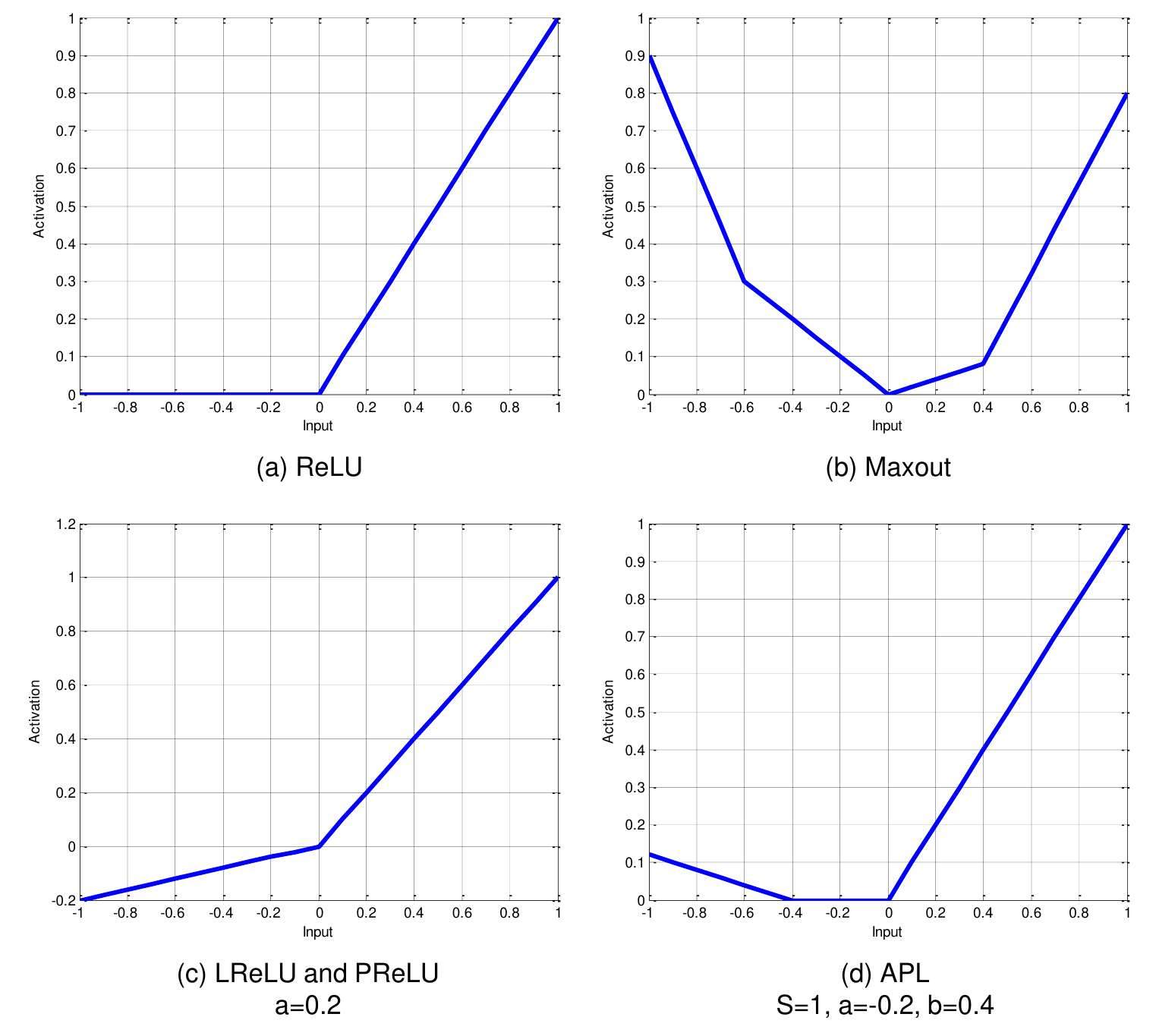}
\caption{Piecewise linear activation functions: ReLU, LReLU, PReLU, APL and
maxout.}
\label{af}
\end{figure} In this section, we first review some activation units
including ReLU, LReLU, PReLU, APL and maxout. Then we introduce two basic laws
in psychophysics and neural sciences: Webner-Fechner law~\cite{fechner} and
Stevens law~\cite{stevens}, as well as our motivation.
\subsection{Rectified Units}
\begin{itemize}
\item Rectified Linear Unit (ReLU) and Its Generalizations\\ ReLU~\cite{ReLU} is defined as
\begin{equation}
\label{eq:ReLU} h(x_i) = \max (0,x_i )
\end{equation} 
where $x_i$ is the input and $h(x_i)$ is the output. LReLU ~\cite{LReLU} assigns a slope to its negative input. It is defined as
\begin{equation}
\label{eq:LReLU} h(x_i) = \min (0,a_i x_i ) + \max (0,x_i )
\end{equation}
where $a_i\in(0,1)$ is a predefined slope. PReLU is only different from LReLU in that the former needs to learn the slope parameter
$a_i$ via back propagation during the training phase.


\item Adaptive Piecewise Linear Units (APL) 

APL is defined as a sum of
hinge-shared functions:
\begin{equation}
  \label{eq:APL} h(x_i ) = \max (0,x_i ) + \sum\limits_s^S {a_i^s \max (0, - x_i
+ b_i^s )},
\end{equation} where $S$ is the number of hinges, and the variables $a_i^s,b_i^s,
i\in 1,...,S$ are parameters of linear functions. 

\quad One disadvantage of APL is it explicitly forces the rightmost line to have unit slope 1 and bias
0. Although it is stated that if the output of APL
serves as the input to a linear function $wh(x_i)+z$, the linear function will
restore the freedom of the rightmost line which is lost due to the constraint, we argue that
this does not always hold because in many cases for deep networks, the function taking the output of APL as the input is non-linear or unrestorable,
such as local response normalizatioin~\cite{krizhevsky2012imagenet} and
dropout~\cite{krizhevsky2012imagenet}.

\item Maxout Unit\\ Maxout unit takes as the input the output of multiple linear
functions and returns the largest:
\begin{equation}
  \label{eq:5} h(x_i ) = \mathop {\max }\limits_{k \in \{ 1,...,K\} } w^k \cdot
x_i + b^k.
\end{equation}
\end{itemize} 
In theory, maxout can approximate any convex
function~\cite{maxout}, but unfortunately, it lacks the ability to learn non-convex
functions. Moreover, a large number of extra parameters introduced by the  $K$ linear functions of each
hidden maxout unit result in large storage memory cost
and considerable training time, which  affect the training efficiency of very deep
CNNs, \textit{e.g.} GoogLeNet~\cite{googlenet}.\\

\subsection{Basic Laws in Psychophysics and Neural Sciences} The Webner-Fechner
law~\cite{fechner} and the Stevens law~\cite{stevens} are two basic laws in
psychophysics~\cite{animalphysiology}~\cite{neuralcoding} and neural
sciences~\cite{actionratedayan2001theoretical}. Webner first observed through
experiments that the amount of change needed for sensory detection to occur
increases with the initial intensity of a stimulus, and is proportional to
it~\cite{weber1851annotationes}. Based on Webner's work, Fechner proposed the
Webner-Fechner law which developed the theory by stating that the subjective sense of intensity is related
to the physical intensity of a stimulus by a logarithmic function, which is
formulated as Eqn.~(\ref{eq:Webner-Fechner}) and shown in Figure~\ref{activationlaw}(a). Stevens refuted the Webner-Fechner law by arguing that the
subjective intensity is related to the physical intensity of a stimulus by a power function~\cite{neuralcoding}, which is formulated as Eqn.~(\ref{eq:Stevens}) and shown
in Figure~\ref{activationlaw}(b). The two laws have been verified through lots of experiments~\cite{naturecounting}. For example, in vision, the
amount of change in brightness with respect to the present brightness accords with the
Webner-Fechner law, \textit{i.e.,} Eqn.~(\ref{eq:Stevens}).~\cite{stevens} 
 shows various examples, one of which is that  the perception of  
pain and taste follows the Stevens law but with different exponent values. In neural
sciences, these two laws also explain many properties of sensory
neurons and the response characteristics of receptor
cells~\cite{naturecounting}~\cite{actionratedayan2001theoretical}. The more detailed
discussion is beyond the range of this paper. 

Motivated by the previous research on these two laws, we propose SReLU
which imitates the logarithm function and the power function given by the Webner-Fechner law and the Stevens law, repectively, and uses piecewise linear functions to approximate non-linear convex and non-convex functions. Through  experiments, we find that our method can be universally used for current deep networks, and significantly boosts the performance.

\section{S-shaped Rectified Linear Units (SReLU)} In this section, we introduce
in detail our proposed SReLU. Firstly, we present the definition and the training process of SReLU. Secondly, we propose a method to initialize the parameters of SReLU as a good starting point for training. Finally, we discuss the relationship of SReLU with other activation
functions.\\

\subsection{Definition of SReLU} SReLU is essentially defined as a combination of three
linear functions, which perform mapping $\mathbb{R} \to \mathbb{R}$ with the
following formulation: \\
\begin{equation}
  \label{eq:SReLU} h(x_i) = \left\{ \begin{array}{l} t_i^r + a_i^r (x_i - t_i^r
),\,\,\,x_i \ge t_i^r \\
x_i,\qquad\qquad\, t_i^r > x_i > t_i^l \\
t_i^l + a_i^l (x_i - t_i^l ),\,\,\,\,\,x_i \le t_i^l \\
 \end{array} \right.
\end{equation} 
where $\left\{t_i^r, a_i^r, t_i^l, a_i^l\right\}$ are four
learnable parameters used to model an individual SReLU activation unit. The subscript $i$
indicates that we allow SReLU to vary in different channels. As shown in Figure~\ref{activationlaw}(c)(d), in the positive direction, $a_i^r$ is the slope of the right line when the
inputs exceed the threshold $t_i^r$. Symmetrically, $t_i^l$ is used to represent
another threshold in the negative direction. When the inputs are smaller than $t_i^l$,
the outputs are calculated by the left line. When the inputs of SReLU fall into the range of
$\left( t_i^l, t_i^r \right)$, the outputs are linear functions with unit slope 1 and bias 0.

By designing SReLU in this way, we hope that it can imitate the
formulations of multiple non-linear functions, including the logarithm function (Eqn.~(\ref{eq:Webner-Fechner})) and the power function (Eqn.~(\ref{eq:Stevens})) given by the Webner-Fechner law and the Stevens law, respectively. As shown in Figure~\ref{activationlaw}(c), when $t_i^r >1, a_i^r > 0$, the
positive part of SReLU imitates the power function with
the exponent $e$ larger than 1; when $1 > t_i^r > 0, a_i^r > 0$,
the positive part of SReLU imitates the logarithm function; when $t_i^r=1, a_i^r > 0$, SReLU follows the power function with the exponent 1. For the negative part of SReLU, we have a similar observation except for the inverse representation of the logarithm function and the power function as analyzed for its positive counterpart. The reason for setting the middle line to be a linear function
with slope 1 and bias 0 when the input is within the range $\left( t_i^l, t_i^r \right)$ is that  it can better approximate both Eqn.~(\ref{eq:Webner-Fechner}) and Eqn.~(\ref{eq:Stevens}) using such a function, because the change of the outputs with respect to the inputs is slow when the inputs are in small magnitudes.

Unlike APL which restricts the form of rightmost line, we do not apply any constraints or regularization to the
parameters, thus both the thresh parameters and slope parameters can be learned freely as the training goes on. It is
noteworthy that no divergence of deep networks occurs although SReLU is allowed to be trained without any constraints in all of our experiments. As shown in Table 3, the learned parameters are all in reasonable condition.

In our method, we learn an independent SReLU following each channel of
kernels. Thus the number of the parameters for SReLU in the deep
networks is only $4N$, where $N$ is the overall number of
kernel channels in the whole network. Compared with the large number of parameters in CNNs,
\textit{e.g.} 5 million parameters in GooLeNet~\cite{googlenet}, such an increase in the number
of parameters (21.7K in GoogLeNet with SReLU, as shown in Table~\ref{imagenet_compare}) is negligible. This is a good
property of SReLU, because on one hand, we avoid the overfitting effectively by increasing only a negligible number of parameters, and on the
other hand we keep the memory size and the computing time almost
unchanged. Similar to PReLU~\cite{PReLU}, we also try the channel-shared variant of SReLU. In this case the number of SReLU is equal to the overall number of layers in the deep network. In Tabel 1, we compare the performance of these two variants of SReLU on CIFAR-10 without data augmentation and find that the channel-wise version performs slightly better than the channel-shared version.

\begin{table}[!t]
\caption{Comparison of error rates between the channel-shared variant and
the channel-wise variant of SReLU on CIFAR-10 without data augmentation.}
\centering \small
\begin{tabular}{ll} 
\hline Model & Error Rates \\ 
\hline NIN + ReLU [Lin
et. al.]  & 10.43\% \\ 
NIN + SReLU (channel-shared) & 9.01\% \\ 
NIN + SReLU (channel-wise) \ \ \ \ \ \ \ \ & 8.41\% \\ \hline
\end{tabular} 
\label{channelwise-shared}
\end{table}

With respect to the training of SReLU, we use the gradient descent algorithm and
jointly train the parameters of SReLU with the deep networks. The update rule of
$\left\{t_r, a_r, t_l, a_l\right\}$ is derived by the chain rule:
\begin{equation}
  \label{eq:chain-rule} \frac{{\partial L}}{{\partial t_i^r }} =
\sum\limits_{x_i } {\frac{{\partial L}}{{\partial h(x_i )}}} \frac{{\partial
h(x_i )}}{{\partial o_i }}
\end{equation} 
where $o_i\in\left\{t_i^r ,a_i^r ,t_i^l ,a_i^l\right\}$ and $L$
represents the objective function of the deep network. The term
${\frac{{\partial L}}{{\partial h(x_i )}}}$ is the gradient back-propagated from the
higher layer of SReLU. The summation $\sum\nolimits_{x_i } $ is applied in all positions of
the feature map. For the channel-shared variant, the gradient of $o_i$ is
$\frac{{\partial L}}{{\partial o_i }} = \sum\nolimits_i {\sum\nolimits_{x_i }
{\frac{{\partial L}}{{\partial h(x_i )}}} } \frac{{\partial h(x_i )}}{{\partial
o_i }}$, where $\sum\nolimits_{x_i } $ is the sum over all channels in each
layer. Specifically, the gradient for each parameter of SReLU is given by
\begin{equation}
  \label{eq:gradient-srelu}
\begin{array}{l} \frac{{\partial h(x_i )}}{{\partial t_i^r }} = I\{ x_i \ge
t_i^r \} (1 - a_i^r) \\ \frac{{\partial h(x_i )}}{{\partial a_i^r }} = I\{ x_i
\ge t_i^r \} (x_i - t_i^r ) \\ \frac{{\partial h(x_i )}}{{\partial t_i^l }} =
I\{ x_i \le t_i^l \} (1 - a_i^l) \\ \frac{{\partial h(x_i )}}{{\partial a_i^l }}
= I\{ x_i \le t_i^l \} (x_i - t_i^l ) \\
 \end{array}
\end{equation} where $I\{ \cdot \} $ is an indicator function and $I\{ \cdot
\}=1 $ when the expression inside holds true, otherwise $I\{ \cdot \}=0 $. By
this way, the gradient of the input is
\begin{equation}
  \label{eq:gradient-input} h(x_i ) = \left\{ \begin{array}{l} a_i^r ,\,\,\,\,x_i
\ge t_i^r \\ 1\,\,\,,\,\,\,\,t_i^r  > x_i > t_i^l \\ 
a_i^l ,\,\,\,\,x_i \le t_i^l .\\
 \end{array} \right.
\end{equation}

The rule for updating $o_i$ by momentum method is:
\begin{equation}
  \label{eq:update-srelu} \Delta o_i : = \mu \Delta o_i + \varepsilon
\frac{{\partial L}}{{\partial o_i }}.
\end{equation}
Here $\mu$ is the momentum and $\varepsilon$ is the learning
rate. Because the weight decay term tends to pull the parameters to zero, we do
not use weight decay ($l_2$ regularization) for $o_i$.\\

\subsection{Adaptive Initialization of SReLU}
\begin{figure}[!t] \centering 
\includegraphics[width=6cm]{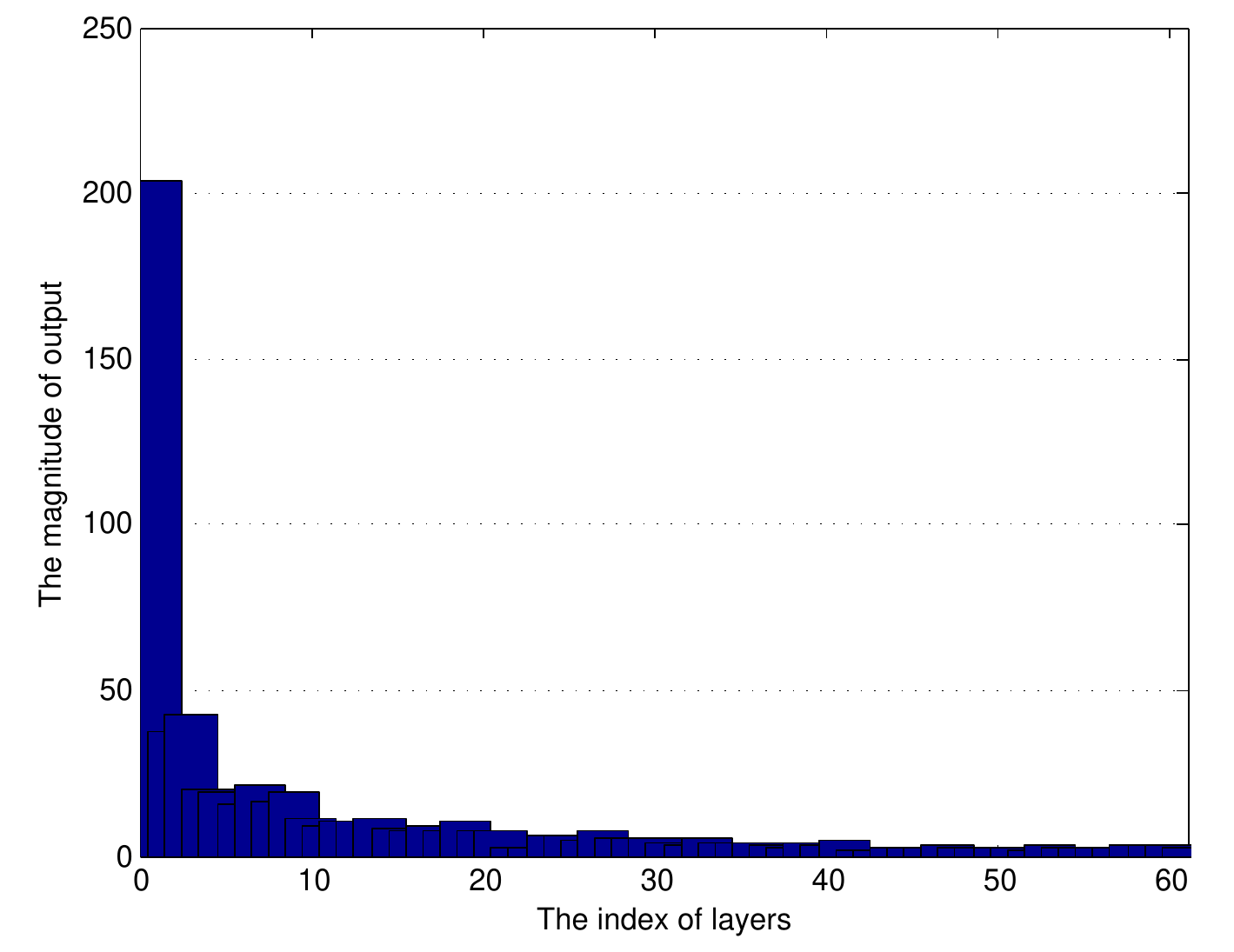} 
\caption{The distribution of the magnitude of the input to SReLU following convolution layers in GoogLeNet. The indexes of convolution layers follow a low-level to high-level order. The magnitudes shown here are calculated by averaging the activations of all SReLUs in each layer.}
\label{magnitude_googlenet}
\end{figure} One problem we are faced in training SReLU is how to
initialize the parameters in SReLU. An intuitive way is to set the parameters manually. However,
such an initialization method is cumbersome. Furthermore, if the manually set initialization values are not appropriate, \textit{e.g.} too large or too small compared with the real value of its input, SReLU may not work well. For example, if $t_i^r$ is set to be very large, based on Eqn.~(\ref{eq:gradient-srelu}), nearly all the inputs for the SReLU will lie in the
left part of $t_i^r$, which will cause $t_i^r$ and $a_i^r$ to be insufficiently learned. In current deep networks, the magnitude of the inputs in
each layer varies a lot (see Figure~\ref{magnitude_googlenet}), making it more
difficult to manually set parameters. To deal with this problem, we propose to
firstly initialize each $o_i$ to be $\{\tilde t_i , 1, 0, \tilde a_i \}$ in all
layers, where $\tilde t_i$ is any positive real number and $\tilde a_i \in
\left(0, 1\right)$, and we ``freeze" the update of the parameters of SReLU during the initial several training epochs. By this method, SReLU is degenerated into a conventional LReLU at
the beginning of the training. Then upon the end of the ``freezing" phase, we set $t_i^r$ to be the largest $k^{th}$ value of each SReLU's input from all training data, \textit{i.e.},
\begin{equation}
  \label{eq:initialization-srelu} t_i^r=\textrm{supp}\left(\mathbf{X_i},k\right)
\end{equation} 
where $\textrm{supp}\left(\mathbf{X}, k\right)$ calculates the $k^{th}$
largest value from the set $\textbf{X}$, and $\mathbf{X_i}$ represents all the input
values of an individual SReLU. Our initialization method offers following two advantages.
Firstly, it learns adaptively the initial values of $t_i$ to fit better to the real distributions of the training data, thus providing a good starting point for the training of SReLU. Secondly, it enables SReLU to re-use the per-trained model with LReLU, thus it can reduce the training time compared with training the whole network from the scratch.

\subsection{Comparison with Other Activation Functions} In this part, we compare
our method with five published nonlinear activation functions: ReLU, LReLU,
PReLU, APL and maxout.

By checking Eqn.~(\ref{eq:ReLU}), Eqn.~(\ref{eq:LReLU}) and Eqn.~(\ref{eq:SReLU}), it can be easily concluded that ReLU, LReLU and PReLU can be seen as special cases of
SReLU. Specifically, when $t_i^r\geq 0,a_i^r=1,t_i^l=0,a_i^l=0$, SReLU
is degenerated into ReLU; when $t_i^r \geq 0, a_i^r=1, t_i^l=0, a_i^l > 0$,
SReLU is transformed to LReLU and PReLU. However, ReLU, LReLU and PReLU can
only approximate convex functions, while SReLU is able to approximate both
convex and non-convex functions. Compared with APL, when the inputs have large magnitudes and lie in the rightmost region of the activate function, SReLU allows its parameters to take more flexible values and gives output features with adaptive scaling over the inputs. This is similar to the Webner-Fechner law that has logarithm function form to suppress the outputs for the input with too large magnitude. SReLU models such suppression effect by learning the slope of its rightmost line adaptively. In contrast, APL constrains the output to be same as input even when the inputs have very large magnitudes. This is the key difference between SReLU and APL and also the main reason why SReLU consistently outperforms APL. The experimental results shown in Table 2 clearly demonstrate this point. Without data augmentation and the proposed initialization strategy, NIN + SReLU outperforms NIN + APL by 0.98\% and 3.04\% on CIFAR-10 and CIFAR-100, respectively. Compared to maxout, which can only approximate convex functions and introduces a large number of extra parameters, SReLU needs much less parameters, therefore SReLU is more suitable for training very deep networks, \textit{e.g.} GoogLeNet.

\section{Experiments and Analysis}
\subsection{Overall Settings} To evaluate our method thoroughly, we conduct
experiments on four datasets with different scales, including CIFAR-10, CIFAR-100~\cite{cifar}, MNIST~\cite{MNIST} and a much larger dataset, ImageNet~\cite{imagenet} with two popular deep networks, \textit{i.e.,} NIN~\cite{NIN} and GoogLeNet~\cite{googlenet}.
NIN is used on CIFAR-10, CIFAR-100 and MNIST and GoogLeNet
is used on ImageNet. NIN replaces the single linear
convolution layers in the conventional CNNs by multilayer perceptrons, and uses
the global average pooling layer to generate
feature maps for each category. Compared to NIN, GoogLeNet is much larger with
22 layers built on \textit{Inception} model, which can be seen as a deeper and
wider extension of NIN. Both these two networks have achieved state-of-the-art
performance on the datasets we use.

Since we mainly focus on testing the effects of SReLU on the performance of deep
networks, in all our experiments, we only replace the ReLU in the original networks with SReLU and keep the other parts of
networks unchanged. For the setting of hyperparameters (such as learning rate,
weight decay and dropout ratio, etc.), we follow the published
configurations of original networks. To compare LReLU with our method, we try
different slope values in Eqn.~(\ref{eq:LReLU}) and picks the one that gets the best performance on validation set. For
PReLU in our experiments, we follow the initialization methods
presented in~\cite{PReLU}. For every dataset, we randomly sample 20\% of the
total training data as the validation set to configure the needed hyperparameters in different methods. After fixing
hyperparameters, we train the model from the scratch with the whole
training data. For SReLU, we use $a^l=0.2$ and $k=\lceil0.9\times|\mathbf{X_i}|\rceil$ for all datasets. In
all experiments, we \emph{ONLY} use single model and single view test.

We choose Caffe~\cite{caffe} as the platform to conduct our experiments. To
reduce the training time, four NVIDIA TITAN GPUs are employed in parallel for
training. Other hardware information of the PCs we use includes Intel Core i7
3.3GHz CPU, 64G RAM and 2T hard disk. The codes of SReLU are available at https://github.com/AIROBOTAI/caffe/tree/SReLU.
\begin{table}[!t]
	\scriptsize
	\caption{Error rates on CIFAR-10 and CIFAR-100. In the column for comparing the
		no. of parameters, the number after ``+" is the extra number of parameters (in KB)
		introduced by corresponding methods. For the row of NIN + APL, 5.68K and 2.84K
		correspond to the extra parameters for CIFAR-10 and CIFAR-100,
		respectively.} \centering
	\begin{tabular}{llcc} \hline Model & No. of Param.(MB) & CIFAR-10 & CIFAR-100
		\\ \hline \multicolumn{4}{c}{ Without Data Augmentation} \\
		\hline  Maxout & \textgreater 5M & 11.68\% &38.57\% \\
		Prob maxout & \textgreater 5M & 11.35\% & 38.14\% \\
		APL & \textgreater 5M & 11.38\% & 34.54\% \\
		DSN  & 0.97M & 9.78\% & 34.57\% \\
		Tree based priors & -& - & 36.85\% \\
		\hline NIN  & 0.97M & 10.41\% & 35.68\% \\
		NIN + ReLU & 0.97M & 9.67\% & 35.96\% \\
		NIN + LReLU  & 0.97M & 9.75\% & 36.00\% \\
		NIN + PReLU(ours) & 0.97M + 1.42K & 9.74\% & 35.95\% \\
		NIN + APL & 0.97M + 5.68K/2.84K & 9.59\% & 34.40\% \\
		NIN + SReLU\footnotemark(ours) & 0.97M + 5.68K & 8.61\% & 31.36\% \\
		NIN + SReLU (ours)& 0.97M + 5.68K & \textbf{8.41\%} & \textbf{31.10\%} \\
		\hline \multicolumn{4}{c}{ With Data Augmentation} \\
		\hline Maxout & \textgreater 5M & 9.38\% &- \\
		Prob maxout  & \textgreater 5M & 9.39\% & - \\
		APL& \textgreater 5M & 9.89\% & 33.88\% \\
		DSN & 0.97M & 8.22\% & - \\
		\hline NIN & 0.97M & 8.81\% & - \\
		NIN + ReLU & 0.97M & 7.73\% & 32.75\% \\
		NIN + LReLU & 0.97M & 7.69\% & 32.70\% \\
		NIN + PReLU (ours) & 0.97M + 1.42K & 7.68\% & 32.67\% \\
		NIN + APL & 0.97M + 5.68K/2.84K & 7.51\% & 30.83\% \\
		NIN + SReLU (ours) & 0.97M + 5.68K & \textbf{6.98\%} & \textbf{29.91\%} \\
		\hline
	\end{tabular}
	\label{table:CIFARRESULT}
\end{table}\footnotetext{Manually set initialization parameters in SReLU}

\subsection{CIFAR} The CIFAR-10 and CIFAR-100 datasets contain
color images with size of 32x32 from 10 and 100 classes, respectively. Both of them
have 50,000 training images and 10,000 testing images. The preprocessing methods
follow the way used in~\cite{maxout}.  The comparison results of SReLU with
other methods (including maxout~\cite{maxout}, prob maxout~\cite{probmaxout}, APL~\cite{learnactivationfunction}, DSN~\cite{dsn}, tree based priors~\cite{treebased}, NIN~\cite{NIN}, etc.) on these two datasets either when the data augmentation is applied
or not are shown in Tabel~\ref{table:CIFARRESULT}, from which we can see that our proposed SReLU
achieves the best performance against all the compared methods.

When no data augmentation is used, compared with ReLU, LReLU and PReLU, our
method reduces the error significantly by 1.26\%, 1.34\%, 1.33\% on
CIFAR-10, respectively. On CIFAR-100, the error reduction is 4.86\%, 4.90\%, 4.85\%,
respectively. SReLU also demonstrates superiority by surpassing other activation
functions including APL and maxout. When compared with other deep network
methods, such as tree based priors and DSN, our method also beats them
by a remarkable gap, demonstrating a promising ability to help boost the
performance of deep models. We also compare the number of parameters used in
each method, from which we notice that SReLU only incurs a  very slight increase
(5.68K) to the total number of parameters (0.97M in original NIN). APL uses the
same number of additional parameters as SReLU on CIFAR-10, but its performance in
either case of applying data augmentation or not is inferior to our method. The convergence curve of SReLU with other methods on CIFAR-10 and CIFAR-100 are shown in Figure 4(a) and Figure 4(b), respectively.

To observe the learned parameters of SReLU, we list in Table~\ref{showSReLUparam} the parameters' values after the training phase. Since the 
SReLUs we use are channel-wise, we simply calculate the average of the input for 
all SReLUs in the same layer. It is interesting to observe that SReLUs in different
layers  learn meaningful parameters in coincide with our motivations. For example, the SReLUs following conv1 and cccp1  learns $a^r$ less
than 1 (0.81 and 0.77, respectively) on CIFAR-10, while SReLUs following conv3 and cccp5 on CIFAR-100 learns $a^r$ larger than 1 (1.42 and 1.36, respectively). SReLU following conv2 on CIFAR-10 learns $a^r$ nearly equal to 1 (1.01). These experimental results verify that SReLU has a strong ability to learn various forms of nonlinear functions, which can either
be convex or non-convex. Moreover, in Table 3, we can see that $t^r$ is of very large value in higher layers. It's because that the inputs of SReLU have higher average values than the ones in lower layers. Therefore, SReLU in higher layers learns larger $t^r$ for adapting to inputs. This demonstrates the strong adaptive ability of SReLU to distribution of its inputs.

In the experiments on the augmented version of CIFAR-10 and CIFAR-100, we simply use random horizontal reflection during training for both datasets. In this case, SReLU still consistently outperforms other methods.
\begin{table}[]
	\centering 
	\caption{The parameters' values of SReLU after training with NIN on CIFAR-10 and CIFAR-100, respectively. The layers listed in the tabel are all convolution layers. ``conv" layers are with kernel sizes larger than 1 and ``cccp" layers are with kernel sizes equal to 1. Each layer in the tabel is followed by channel-wise SReLUs. For more details of the layers in NIN, please refer to~\cite{NIN}}
	\label{my-label}\scriptsize
	\begin{tabular}{c|c|c|c|c}
		\hline
		\multirow{2}{*}{layers} & \multicolumn{4}{c}{CIFAR-10 / CIFAR-100} \\ \cline{2-5}
	             	& $t^r$            & $t^l$          & $a^r$         & $a^l$  \\ \hline
		conv1  & 0.91  / 0.73    & -0.48 / -0.68   & 0.81 / 0.62   & -0.25 / -0.22\\ \hline
		cccp1  & 1.06  / 0.52    & -0.36 / -0.34   & 0.77 / 0.38   & -0.04 / 0.04 \\ \hline
		cccp2  & 1.27  / 0.37     & -0.20 / -0.26  & 0.47 / 0.51  & 0.39 / 0.44\\ \hline
		conv2  & 5.32  / 4.02    & -0.31 / -0.51   & 1.01 / 0.88  & 0.07 / 0.06\\ \hline
		cccp3  & 6.95  / 4.73    & -0.21 / -0.79  & 0.92 / 0.64  & -0.01 / 0.05\\ \hline
		cccp4  & 8.18  / 5.79    & -0.08 / -0.13  & 0.77 / 0.56  & 0.61  / 0.45\\ \hline
		conv3  & 25.17 / 23.72    & -0.15 / -0.61   & 1.21 / 1.42  & 0.05 / 0.07\\ \hline
		cccp5  & 31.09 / 36.44   & -0.47 / -0.46  & 0.97 / 1.36   & -0.16 / -0.02\\ \hline
		cccp6  & 72.03 / 66.13   & -0.13 / -0.21  & 1.53 / 1.23  & -0.44 / -0.35\\ \hline	
	\end{tabular}
	\label{showSReLUparam} 
\end{table}
\begin{table}[!t]
	\label{MNIST_result} \small
	\caption{Error rates on MNIST without data augmentation. }
    \centering
	\begin{tabular}{lll} \hline Model & No. of Param.(MB) & {Error Rates} \\
		\hline Stochastic Pooling & - & 0.47\% \\
		Maxout& 0.42M & 0.47\% \\
		DSN & 0.35M & 0.35\% \\
		\hline NIN + ReLU & 0.35M & 0.47\% \\
		NIN + LReLU (ours) & 0.35M & 0.42\% \\
		NIN + PReLU (ours) & 0.35M + 1.42K & 0.41\% \\
		NIN + SReLU (ours)& 0.35M + 5.68K & \textbf{0.35\%} \\
		\hline
	\end{tabular}
\end{table}
\subsection{MNIST} MNIST~\cite{MNIST} contains 70,000 28x28 gray scale images of
numerical digits from 0 to 9, divided as 60,000 images for training and 10,000
images for testing.  In this dataset, we do not apply any preprocessing to the
data and only compare models without data augmentation. The experiment results on
this dataset are shown in Tabel 4, from which we see SReLU performs
better than other methods.

\begin{table}[!t] \scriptsize \renewcommand{\arraystretch}{1}
	\caption{Error rates on ImageNet. Tests are by single model single view.}
	\label{table_example} \centering
	\begin{tabular}{lll} \hline Model & No. of Param.(MB) & Error Rates \\ \hline
		GoogLeNet\footnotemark & 5M & 11.1\% \\ GoogLeNet + SReLU (ours)& 5M +
		21.6K & \textbf{9.86\%} \\ \hline
	\end{tabular}
	\label{imagenet_compare}
\end{table}
\footnotetext{\scriptsize{$\textrm{https://}\textrm{github}.\textrm{com/}\textrm{BVLC/}\textrm{caffe/}\textrm{tree/}\textrm{master/}\textrm{models/}\textrm{bvlc}_{\textrm{-}}\textrm{googlenet}$}}

\subsection{ImageNet} To further evaluate our method on large-scale datasets,
we perform a much more challenging image classification task on
1000-class ImageNet dataset, which contains about 1.2 million training images, 50,000
validation images and 100,000 test images. Our baseline model is GoogLeNet
model, which achieved the best performance on image classification in ILSVRC 2014~\cite{ILSVRC15}. We run experiments using the publicly available configurations in
Caffe~\cite{caffe}. For this dataset, no additional preprocessing method is used
except subtracting the image mean from each input raw image.

Table~\ref{imagenet_compare} compares the performance of GoogLeNet using SReLU and the
original GoogLeNet released by Caffe. The GoogLeNet with SReLU achieves
significant improvement (1.24\%) on this challenging dataset compared with the original
GoogLeNet using ReLU, at the cost of only 21.6K additional parameters (versus
the total number of  5M parameters in the original GoogLeNet).

\begin{figure}[!t] \centering \hspace{-0.2cm}
	\includegraphics[width=8.5cm]{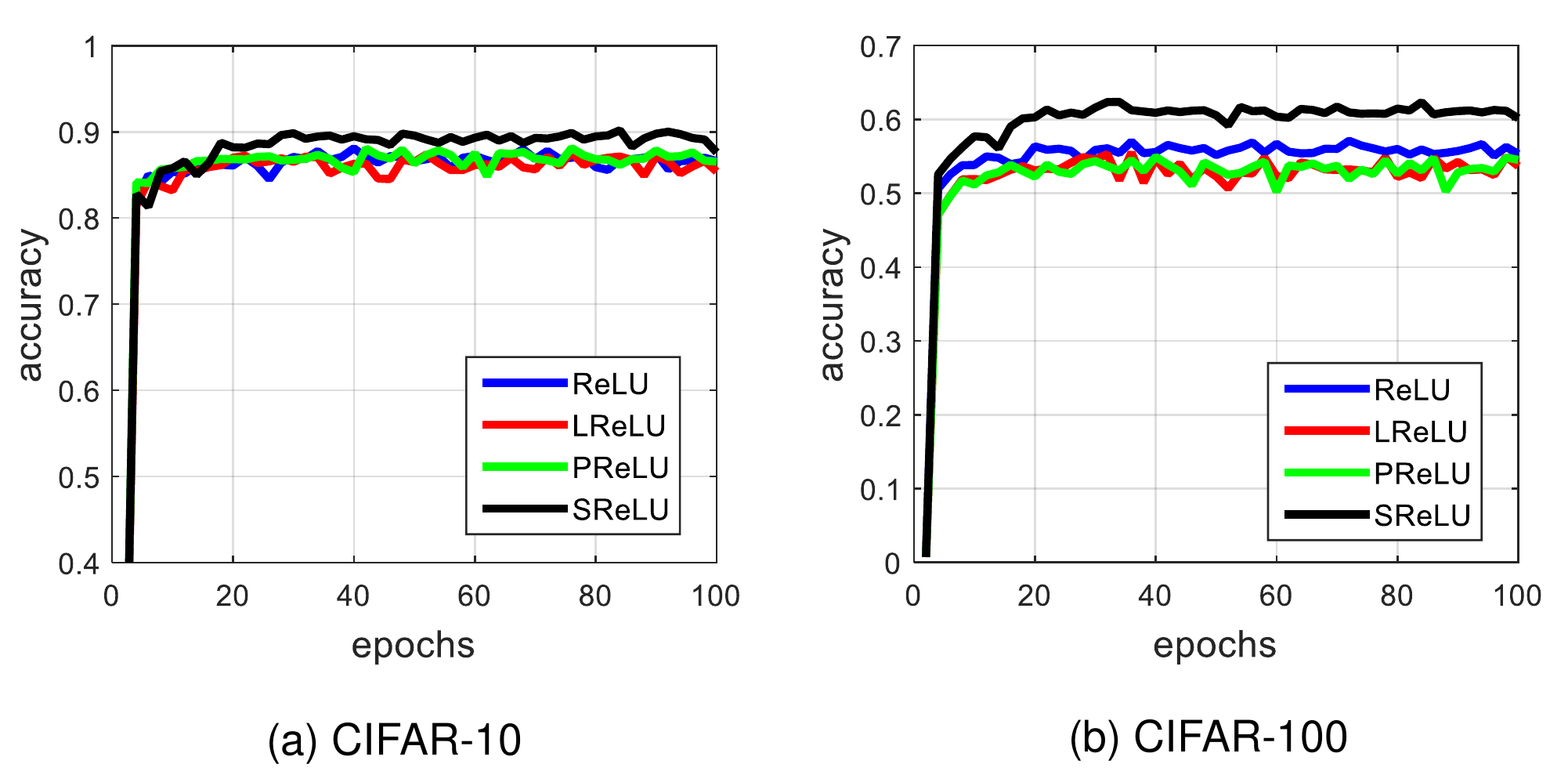}
	\caption{(a) The convergence curves of SReLU and other methods on CIFAR-10. (b) the convergence curves of SReLU and other methods on CIFAR-100.}
	\label{CIFAR-10} 
\end{figure}
\section{Conclusion} In this paper, inspired by the fundamental laws in
psychophysics and neural sciences, we proposed a novel S-shaped rectified linear
unit (SReLU) to be used in deep networks. Compared to other activation functions,
SReLU is able to learn both convex and non-convex functions, and can be universally used in existing deep networks. Experiments on four datasets including
CIFAR-10, CIFAR-100, MNIST and ImageNet with NIN and GoogLeNet demonstrate
that SReLU effectively boosts the performance of deep networks. In our future work, we will exploit the applications of SReLU in other domains beyond vision, such as NLP.
 \bibliographystyle{aaai} \bibliography{mybibfile}

\begin{thebibliography}{}

\bibitem[\protect\citeauthoryear{Agostinelli \bgroup et al\mbox.\egroup
  }{2014}]{learnactivationfunction}
Agostinelli, F.; Hoffman, M.; Sadowski, P.~J.; and Baldi, P.
\newblock 2014.
\newblock Learning activation functions to improve deep neural networks.
\newblock {\em CoRR} abs/1412.6830.

\bibitem[\protect\citeauthoryear{Cire{\c{s}}an \bgroup et al\mbox.\egroup
  }{2011}]{cirecsan2011convolutional}
Cire{\c{s}}an, D.~C.; Meier, U.; Gambardella, L.~M.; and Schmidhuber, J.
\newblock 2011.
\newblock Convolutional neural network committees for handwritten character
  classification.
\newblock In {\em Document Analysis and Recognition (ICDAR), 2011 International
  Conference on},  1135--1139.
\newblock IEEE.

\bibitem[\protect\citeauthoryear{Dayan and
  Abbott}{2001}]{actionratedayan2001theoretical}
Dayan, P., and Abbott, L.~F.
\newblock 2001.
\newblock {\em Theoretical neuroscience}, volume 806.
\newblock Cambridge, MA: MIT Press.

\bibitem[\protect\citeauthoryear{Deng \bgroup et al\mbox.\egroup
  }{2009}]{imagenet}
Deng, J.; Dong, W.; Socher, R.; Li, L.-J.; Li, K.; and Fei-Fei, L.
\newblock 2009.
\newblock Imagenet: A large-scale hierarchical image database.
\newblock In {\em Computer Vision and Pattern Recognition, 2009. CVPR 2009.
  IEEE Conference on},  248--255.
\newblock IEEE.

\bibitem[\protect\citeauthoryear{Fechner}{1965}]{fechner}
Fechner, G.
\newblock 1965.
\newblock Elements of psychophysics.

\bibitem[\protect\citeauthoryear{Girshick \bgroup et al\mbox.\egroup
  }{2013}]{RCNN}
Girshick, R.~B.; Donahue, J.; Darrell, T.; and Malik, J.
\newblock 2013.
\newblock Rich feature hierarchies for accurate object detection and semantic
  segmentation.
\newblock {\em CoRR} abs/1311.2524.

\bibitem[\protect\citeauthoryear{Goodfellow \bgroup et al\mbox.\egroup
  }{2013}]{maxout}
Goodfellow, I.~J.; Warde-Farley, D.; Mirza, M.; Courville, A.; and Bengio, Y.
\newblock 2013.
\newblock Maxout networks.
\newblock {\em arXiv preprint arXiv:1302.4389}.

\bibitem[\protect\citeauthoryear{He \bgroup et al\mbox.\egroup }{2015}]{PReLU}
He, K.; Zhang, X.; Ren, S.; and Sun, J.
\newblock 2015.
\newblock Delving deep into rectifiers: Surpassing human-level performance on
  imagenet classification.
\newblock {\em CoRR} abs/1502.01852.

\bibitem[\protect\citeauthoryear{Jia \bgroup et al\mbox.\egroup }{2014}]{caffe}
Jia, Y.; Shelhamer, E.; Donahue, J.; Karayev, S.; Long, J.; Girshick, R.~B.;
  Guadarrama, S.; and Darrell, T.
\newblock 2014.
\newblock Caffe: Convolutional architecture for fast feature embedding.
\newblock {\em CoRR} abs/1408.5093.

\bibitem[\protect\citeauthoryear{Johnson, Hsiao, and
  Yoshioka}{2002}]{neuralcoding}
Johnson, K.~O.; Hsiao, S.~S.; and Yoshioka, T.
\newblock 2002.
\newblock Book review: neural coding and the basic law of psychophysics.
\newblock {\em The Neuroscientist} 8(2):111--121.

\bibitem[\protect\citeauthoryear{Krizhevsky and Hinton}{2009}]{cifar}
Krizhevsky, A., and Hinton, G.
\newblock 2009.
\newblock Learning multiple layers of features from tiny images.

\bibitem[\protect\citeauthoryear{Krizhevsky, Sutskever, and
  Hinton}{2012}]{krizhevsky2012imagenet}
Krizhevsky, A.; Sutskever, I.; and Hinton, G.~E.
\newblock 2012.
\newblock Imagenet classification with deep convolutional neural networks.
\newblock In {\em Advances in neural information processing systems},
  1097--1105.

\bibitem[\protect\citeauthoryear{LeCun \bgroup et al\mbox.\egroup
  }{1998}]{MNIST}
LeCun, Y.; Bottou, L.; Bengio, Y.; and Haffner, P.
\newblock 1998.
\newblock Gradient-based learning applied to document recognition.
\newblock {\em Proceedings of the IEEE} 86(11):2278--2324.

\bibitem[\protect\citeauthoryear{Lee \bgroup et al\mbox.\egroup }{2014}]{dsn}
Lee, C.-Y.; Xie, S.; Gallagher, P.; Zhang, Z.; and Tu, Z.
\newblock 2014.
\newblock Deeply-supervised nets.
\newblock {\em arXiv preprint arXiv:1409.5185}.

\bibitem[\protect\citeauthoryear{Lin, Chen, and Yan}{2013}]{NIN}
Lin, M.; Chen, Q.; and Yan, S.
\newblock 2013.
\newblock Network in network.
\newblock {\em CoRR} abs/1312.4400.

\bibitem[\protect\citeauthoryear{Maas, Hannun, and Ng}{2013}]{LReLU}
Maas, A.~L.; Hannun, A.~Y.; and Ng, A.~Y.
\newblock 2013.
\newblock Rectifier nonlinearities improve neural network acoustic models.
\newblock In {\em Proc. ICML}, volume~30.

\bibitem[\protect\citeauthoryear{Nair and Hinton}{2010}]{ReLU}
Nair, V., and Hinton, G.~E.
\newblock 2010.
\newblock Rectified linear units improve restricted boltzmann machines.
\newblock In {\em Proceedings of the 27th International Conference on Machine
  Learning (ICML-10)},  807--814.

\bibitem[\protect\citeauthoryear{Nieder}{2005}]{naturecounting}
Nieder, A.
\newblock 2005.
\newblock Counting on neurons: the neurobiology of numerical competence.
\newblock {\em Nature Reviews Neuroscience} 6(3):177--190.

\bibitem[\protect\citeauthoryear{Randall \bgroup et al\mbox.\egroup
  }{2002}]{animalphysiology}
Randall, D.; Burggren, W.~W.; French, K.; and Eckert, R.
\newblock 2002.
\newblock {\em Eckert animal physiology}.
\newblock Macmillan.

\bibitem[\protect\citeauthoryear{Russakovsky \bgroup et al\mbox.\egroup
  }{2015}]{ILSVRC15}
Russakovsky, O.; Deng, J.; Su, H.; Krause, J.; Satheesh, S.; Ma, S.; Huang, Z.;
  Karpathy, A.; Khosla, A.; Bernstein, M.; Berg, A.~C.; and Fei-Fei, L.
\newblock 2015.
\newblock {ImageNet Large Scale Visual Recognition Challenge}.
\newblock {\em International Journal of Computer Vision (IJCV)}  1--42.

\bibitem[\protect\citeauthoryear{Springenberg and
  Riedmiller}{2013}]{probmaxout}
Springenberg, J.~T., and Riedmiller, M.
\newblock 2013.
\newblock Improving deep neural networks with probabilistic maxout units.
\newblock {\em arXiv preprint arXiv:1312.6116}.

\bibitem[\protect\citeauthoryear{Srivastava and
  Salakhutdinov}{2013}]{treebased}
Srivastava, N., and Salakhutdinov, R.~R.
\newblock 2013.
\newblock Discriminative transfer learning with tree-based priors.
\newblock In Burges, C.; Bottou, L.; Welling, M.; Ghahramani, Z.; and
  Weinberger, K., eds., {\em Advances in Neural Information Processing Systems
  26}. Curran Associates, Inc.
\newblock  2094--2102.

\bibitem[\protect\citeauthoryear{Stevens}{1957}]{stevens}
Stevens, S.~S.
\newblock 1957.
\newblock On the psychophysical law.
\newblock {\em Psychological review} 64(3):153.

\bibitem[\protect\citeauthoryear{Szegedy \bgroup et al\mbox.\egroup
  }{2014}]{googlenet}
Szegedy, C.; Liu, W.; Jia, Y.; Sermanet, P.; Reed, S.; Anguelov, D.; Erhan, D.;
  Vanhoucke, V.; and Rabinovich, A.
\newblock 2014.
\newblock Going deeper with convolutions.
\newblock {\em arXiv preprint arXiv:1409.4842}.

\bibitem[\protect\citeauthoryear{Weber}{1851}]{weber1851annotationes}
Weber, E.
\newblock 1851.
\newblock Annotationes anatomicae et physiologicae [anatomical and
  physiological annotations].
\newblock {\em Leipzig: CF Koehler}.

\end{thebibliography}
\end{document}